\documentclass{article}
\usepackage{spconf,amsmath,graphicx}
\usepackage{color}
\usepackage{amsfonts}       
\usepackage{nicefrac}       
\usepackage{microtype}      
\usepackage{graphicx}
\usepackage{caption}
\usepackage{subcaption}
\usepackage{bm}
\usepackage{bbm}
\usepackage{multirow}
\usepackage{amssymb}
\usepackage[utf8]{inputenc} 
\usepackage[english]{babel}
\usepackage[T1]{fontenc}    
\usepackage{hyperref}       
\usepackage{url}            
\usepackage{booktabs}       
\usepackage{enumitem}
\graphicspath{ {./images/} }

\providecommand{\keywords}[1]
{
  \small	
  \textbf{\textit{Keywords---}} #1
}

\title{Cross-Domain Sentiment Classification With Contrastive Learning and Mutual Information Maximization}
\twoauthors
 {Tian Li\sthanks{Equal contribution}, Xiang Chen\footnotemark[1]}
	{Peking University\\
	Department of Computer Science}
 {Shanghang Zhang\sthanks{Correspondence author}, Zhen Dong\footnotemark[2], Kurt Keutzer}
	{UC Berkeley\\
	BAIR Lab}

\begin{document}

\maketitle
\begin{abstract}
Contrastive learning (CL) has been successful as a powerful representation learning method. In this work we propose CLIM: Contrastive Learning with mutual Information Maximization, to explore the potential of CL on cross-domain sentiment classification. To the best of our knowledge, CLIM is the first to adopt contrastive learning for natural language processing (NLP) tasks across domains. Due to scarcity of labels on the target domain, we introduce mutual information maximization (MIM) apart from CL to exploit the features that best support the final prediction. Furthermore, MIM is able to maintain a relatively balanced distribution of the model’s prediction, and enlarges the margin between classes on the target domain. The larger margin increases our model's robustness and enables the same classifier to be optimal across domains. Consequently, we achieve new state-of-the-art results on the Amazon-review dataset as well as the airlines dataset, showing the efficacy of our proposed method CLIM. 
\end{abstract}

\section{Introduction}
Domain shift is common in language applications. One is more likely to find "internet" or "PC" in reviews on electronics than on books, while it is more likely to find "writing" or "B.C." in reviews on books than those on electronics. This proposes a fundamental challenge to natural language processing in that many computational models fail to maintain comparable performance across different domains. Formally, a distribution shift happens when a model is trained on data from one distribution (source domain), but the goal is to make good predictions on some other distribution (target domain) that shares the label space with the source. 

In this paper, we study domain adaptation for sentiment classification, i.e. cross-domain sentiment classification, where we have texts with binary sentiment labels ("positive" or "negative") on the source domain, but no labeled texts on the target domain. The goal is to train a model on labeled source data, unlabeled source data, and unlabeled target data, and test its ability to predict the sentiment polarity on the labeled target data. The most prevailing methods in this field aim to learn domain-invariant feature by aligning the source and target domains in the feature space. Pioneering works in this field try to bridge domain gap based on discrepancy. ~\cite{Long2015Learning} first introduces MK-MMD to measure domain discrepancy as the objective function for closing the domain shift. Various works follow this idea by preprocessing the input data to align domains~\cite{sun2015return} or designing MMD-variants as measurements of domain discrepancy~\cite{he2018adaptive}. Another line of work~\cite{ganin2016domainadversarial} introduces a domain classifier and adversarial training to induce domain alignment and learn domain-invariant feature, followed by works combining various neural network models with adversarial training~\cite{clinchant-etal-2016-domain}, and other works that use generative models to enhance adversarial training~\cite{alam-etal-2018-domain}. After the emergence of transformer-based models such as BERT~\cite{devlin2019bert}, one latest work combined adversarial training with BERT to learn enhanced domain-invariant feature, which achieves state-of-the-art results on standard benchmarks. However, both MMD-based approach and adversarial training formulates a minimax problem, which is widely known as hard to optimize~\cite{fedus2017many}. In addition, some recent works~\cite{combes2020domain, li2020rethinking, li2020learning} have discovered that both MMD-based approach and adversarial training implicitly force the distribution matching, which doesn't guarantee good adaptation and can cause inevitable error on target domain under label distribution shift.

Self-supervised representation learning could serve as a good workaround for this problem. Following the classic work structure correspondence learning~\cite{blitzer2006domain}, multiple works ~\cite{yujiang2016learning,ziser-reichart-2018-pivot} have used sentiment-indicating pivot prediction as their auxiliary task for cross-domain sentiment analysis. 
In this work, we adopt contrastive learning~\cite{he2019moco} to extract discriminative feature. Contrastive learning is a method for self-supervised learning, which forces the model to extract similar features from the query and its positive pair, and it utilizes negative samples to form contrast against the queried sample on pretext tasks, in order to learn discriminative representations. Thanks to recent progress~\cite{he2019moco, chen2020simCLR}, contrastive learning has demonstrated its strong power to extract discriminative features. Given that, CERT~\cite{fang2020cert} and DeCLUTR~\cite{giorgi2020declutr} have combined transformer models with contrastive learning to learn discriminative sentence embeddings.

However, contrastive learning alone doesn't necessarily help domain transfer, so we introduce mutual information maximization~\cite{li2020rethinking} to extract discriminative features that can best support the final prediction. With mutual information maximization, our model's prediction distribution will be reasonably balanced and the margin between classes on target domain will be enlarged, thus making our model more robust and enabling the same predictor to be optimal across domains. The intuition is that the decision boundary learnt on the source domain is more likely to fall into the margin on the target domain. Experiments on two standard cross-domain sentiment classification benchmarks show the efficacy of the proposed method, and we also conduct ablation studies to justify that mutual information helps cross-domain sentiment classification under various domain settings. Our main contributions are summarized as follows:

\begin{itemize}
    \item We are the first to introduce contrastive learning to assist domain adaptation on NLP problems.
    \item We propose the model CLIM: \textbf{C}ontrastive \textbf{L}earning with mutual \textbf{I}nformation \textbf{M}aximization to learn discriminative features that best support the final prediction to achieve cross domain adaptation. 
    \item We outperform the state-of-the-art baselines and achieve superior performance on standard benchmarks of cross-domain sentiment classification. 
\end{itemize}

\section{Method}
\subsection{CLIM Framework}
\label{sec:framework}
\begin{figure}
    \centering
    \includegraphics[width=0.45\textwidth]{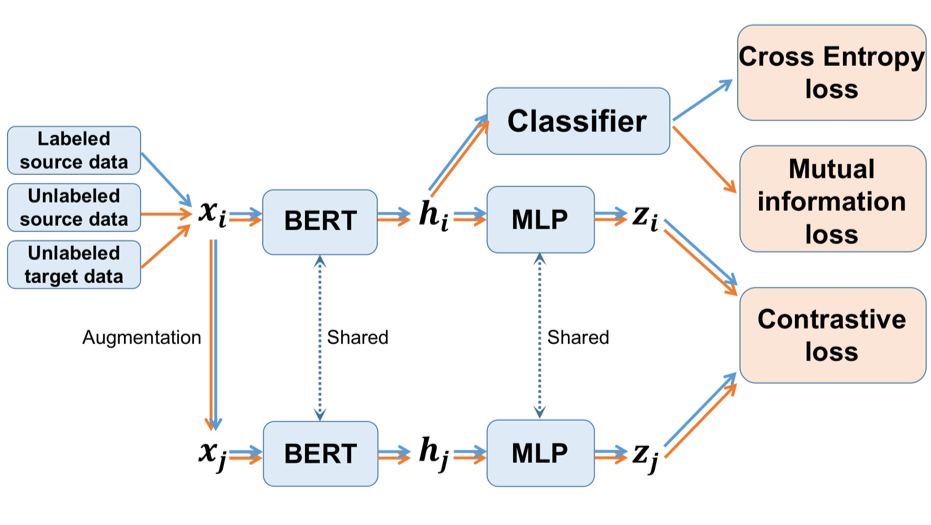}
    \caption{The pipeline of the proposed method. The blue arrows denote the flow of labeled data from the source domain; the orange arrows denote the flow of unlabeled data from both domains. (Best viewed in color.)}
    \label{fig:major_pipeline}
\end{figure}

We propose a framework based on SimCLR~\cite{chen2020simCLR} but elaborate it to fit the domain adaptation problem setting where we have unlabeled data from source and target domains plus labeled data on source domain. As illustrated in Figure {\color{red}\ref{fig:major_pipeline}}, our framework consists of the following components:
\begin{itemize}[leftmargin=*]
    \item \textbf{Positive sample generation:} We use the classic augmentation method, back-translation, to generate positive pair $\bm{x}_j$ for each document $\bm{x}_i$.
    \item \textbf{Feature extractor:} We use pretrained BERT\cite{devlin2019bert} as our feature extractor. 
    Note that the feature extractor is shared among all data, including augmented data. It computes on $\bm{x}_i, \bm{x}_j$ independently and outputs hidden features $\bm{h}_i, \bm{h}_j$.
    \item \textbf{Projection head:} Following SimCLR~\cite{chen2020simCLR}, we adopt an MLP with one hidden layer applying on $\bm{h}_i, \bm{h}_j$ to get the projected representations $\bm{z}_i, \bm{z}_j$ respectively. We find that this MLP projection benefits our model in terms of learning better discriminative features,
    as is also found in SimCLR~\cite{chen2020simCLR}.
    \item \textbf{Contrastive loss:} We apply InfoNCE loss~\cite{chen2020simCLR} to the projected representations $\bm{z}$ for contrastive learning. The contrastive loss function is formulated as:
\begin{equation}
    \mathcal{L}_{con}(\bm{z})=\frac{1}{2N}\sum_{k=1}^{2N}[l(\bm{z}_{2k-1}, \bm{z}_{2k}), l(\bm{z}_{2k}, \bm{z}_{2k-1})]
\end{equation}
    \begin{equation}
        l(\bm{z}_i,\bm{z}_j)=-\log\frac{\exp(\textit{sim}(\bm{z}_i, \bm{z}_j)/\tau)}{\sum_{k=1}^{2N}\mathbbm{1}_{[k\neq i]}\exp(\textit{sim}(\bm{z}_i, \bm{z}_k)/\tau)}
    \end{equation}
    where $\mathbbm{1}_{[k\neq i]}$ is an indicator function equaling to $0$ if $k=i$, and $\textit{sim}(\bm{z}_i, \bm{z}_j)=\bm{z}_i^\top\bm{z}_j/(\|\bm{z}_i\|\|\bm{z}_j\|)$ denotes the cosine similarity between hidden representations $\bm{z}_i$ and $\bm{z}_j$. $N$ is the batch size. $\tau$ is temperature parameter.
    \item \textbf{Sentiment classifier:} We use another one-hidden-layer MLP as sentiment classifier to generate the logits. For the labeled instances, they are used to compute the cross entropy loss w.r.t the ground truth sentiment labels, which formulates our sentiment classification loss $\mathcal{L}_{sent}$.
    \item \textbf{Mutual information loss:} For the unlabeled data, logits from the sentiment classifier are then used to compute the \emph{mutual information loss}. Details of the \emph{mutual information loss} are discussed in Section {\color{red} \ref{sec:mutual_information_loss}}.
    
\end{itemize}

\subsection{Mutual Information Maximization}
\label{sec:mutual_information_loss}
Although contrastive learning is powerful at learning discriminative representations, it doesn't necessarily helps domain adaptation. For example, due to scarcity of labels on the target domain, the model's predictions of data from the target domain may be unreasonably biased to one class. Also, many instances on the target domain may be close to the optimal decision boundary learned on the source domain, thus undermining the robustness of the contrastive learning method. Therefore, we propose to apply mutual information maximization to the unlabeled data from source and target domains. The formulation is $I(X;Y)=H(Y)-H(Y|X)$, where $X$ is the input text and $Y$ is our model's prediction.

Maximizing mutual information can be broken into two parts: maximizing $H(Y)$ and minimizing $H(Y|X)$. The former prevents the prediction of the model from being too biased to one class, so as to maintain the model's prediction distribution on a reasonably balanced status. The latter increases the model's confidence on the prediction, and enlarges the margin between classes. With a larger margin, the decision boundary learned on the source domain is more likely to fall into it, thus allowing the same classifier to be optimal across domains. This can make our model become more robust when testing. Following~\cite{li2020rethinking}, we maximize $H(Y)$ only when it is lower than a pre-defined threshold.

Our mutual information objective function is stated as follows:
\begin{equation}
    \mathcal{L}_{MI}=\mathbb{E}_{y}[\log p_\theta(y)]-\mathbb{E}_{x}[\sum_{y}p_\theta(y|x)\log p_\theta(y|x)]
\end{equation}
where in the first term $y\sim P_\theta(Y)$, and in the second term $x\sim P(X)$. $p_\theta(y|x)$ denotes probability of each sample output from predictor, $P_\theta(Y)$ represents the distribution of predicted target label $Y$. Inspired by ~\cite{li2020rethinking}, we use the average of $p_\theta(y|x)$ in the minibatch as an approximation of $p_\theta(y)$.

\subsection{Joint Learning}
With CLIM, the model is jointly trained on the three objectives. The total loss is formulated as:
\begin{equation}
    \mathcal{L}_{total}=\lambda_{con}\mathcal{L}_{con}+\lambda_{sent}\mathcal{L}_{sent}+\mathcal{L}_{MI}
\end{equation}

\section{Experiments}

\subsection{Experiment Settings}
\label{sec:experiments}
\subsubsection{Cross-domain Sentiment Classification Datasets} 
To evaluate the efficacy of our method CLIM, we implement the domain adaptation framework proposed in {\color{red}\ref{sec:framework}} and name the model as BERT-CLIM. To demonstrate its capability of cross-domain sentiment classification, we conduct experiments on the Amazon-review\footnote{Dataset can be found at http://www.cs.jhu.edu/~mdredze/datasets/sentiment/ index2.html} dataset~\cite{blitzer-etal-2007-biographies} and Airlines dataset\footnote{\leftline{Dataset and process procedures can be found at https://github.com/} quankiquanki/skytrax-reviews-dataset}, following PBLM ~\cite{ziser-reichart-2018-pivot}. As is shown in Table {\color{red}\ref{tab:label_distribution}}, the Amazon-review dataset contains reviews for four kinds of products, books(B), DVD(D), electronics(E), and kitchen(K), each defining a domain. Each domain has 1000 samples labeled as positive and negative respectively (2000 labeled data in total). The number of unlabeled data and its label distribution on each domain is different, as shown in the last two columns of the table. The airlines dataset contains reviews of airlines. We introduce this dataset because it is distant to the domains from the Amazon-review dataset, making it harder for domain adaptation. To align the setting with the Amazon-review dataset, we randomly sample 1000 positive and 1000 negative reviews from the airlines dataset as labeled data, and the rest of data are removed of labels as unlabeled data.


\begin{table}[h]
\centering
\begin{tabular}{ |c | c| c | c | c | } 
\hline
\textbf{Domains} & \textbf{labeled} & \textbf{unlabeled} & \textbf{pos:neg} \\ 
\hline
Books & 2000 & 6000 & 6.43:1\\ 
\hline
DVD & 2000 & 34741 & 7.39:1 \\ 
\hline
Electronics & 2000 & 13153 & 3.65:1 \\
\hline
Kitchen & 2000 & 16785 & 4.61:1 \\
\hline
Airlines & 2000 & 39396 & 1.15:1 \\
\hline
\end{tabular}
\setlength{\abovecaptionskip}{3pt}
\caption{The statistics of domains on the Amazon-review dataset and the airlines dataset reported by~\cite{ziser-reichart-2018-pivot}. "pos:neg" denotes the ratio of unlabeled positive samples over unlabeled negative samples on that domain. }
\label{tab:label_distribution}
\end{table}

Given the computation cost and the scale of the datasets, we pick 8 pairs of domains as source and target respectively to evaluate the proposed method, namely B$\to$E, K$\to$B, K$\to$D, E$\to$K, K$\to$E, B$\to$K, B$\to$A and A$\to$B. In each of the settings, we randomly sample 400 of the source labeled data for development (dev set), and use the rest 1600 source labeled data, along with unlabeled data from both domains, for training. All labeled data from the target domain are used for testing.

\vspace{-3mm}
\subsubsection{Baselines} We compare our model BERT-CLIM with the state-of-the-art model BERT-DAAT~\cite{du2020adversarial} and baselines that we implement by incorporating previous methods to BERT, including:
\begin{itemize}[leftmargin=*]
\setlength{\itemsep}{0pt}
\setlength{\parsep}{0pt}
\setlength{\parskip}{0pt}
    \item \textbf{Base}: BERT trained on the source labeled data and directly test on the target labeled data.
    \item \textbf{DANN}: BERT combined with the popular adversarial training method, DANN~\cite{ganin2016domainadversarial}. 
    \item \textbf{DAAT}: BERT-DAAT proposed by~\cite{du2020adversarial}, the previous state-of-the-art model.
\end{itemize}
Results of two outstanding non-BERT methods PBLM~\cite{ziser-reichart-2018-pivot} and IATN~\cite{zhang2019interactive} are also included. For the sake of simplicity, we call our model CLIM  for short in the rest of the experiments section.



\subsection{Implementation Details}
\subsubsection{Contrastive Learning Strategy}
Unlike contrastive learning of representations like~\cite{chen2020simCLR}, we have data from two domains, so we consider two contrastive learning strategies: both-domain contrastive learning and in-domain contrastive learning. The former combines the source and target domains, and learns by forming contrast on them altogether, while the latter independently learns on the source domain and target domain. We treat the choice of strategy as a hyperparameter.

\subsubsection{Hyperparameter Tuning}
We adopt the pretrained BERT-base-uncased model as the feature extractor for CLIM and our baselines Base and DANN. We use ReLU as the activation function for projection head, domain classifier and sentiment classifier (if applicable). As for training strategy, we train the models for 10 epochs, using the AdamW~\cite{loshchilov2018fixing} optimizer with learning rate $2\times10^{-5}$, linear learning rate scheduler, linear learning rate warm up, warm up steps 0.1 of total training steps. 

Specifically for CLIM, We set the temperature parameter of the infoNCE loss to $0.05$, and L2 weight decay to $0.01$ following ~\cite{fang2020cert} and ~\cite{giorgi2020declutr}. For back-translation, we set the beam parameter to 1. In-domain contrastive learning is used for the 6 experiments on Amazon-review dataset, and both-domain contrastive learning is used for the two involving airlines dataset. 

In the gradient reversal layer of DANN, we define the training progress as $p = \frac{t}{T}$, where $t$ and $T$ are current training step and the maximum training step respectively, and the adaptation rate $\lambda$ is increased following $\lambda=\frac{2}{1+exp(-\gamma p)}-1$. We choose $\gamma$ from ${0.25, 0.5, 0.75, 1.0}$ on different domain settings. Note that we find with experiments that $\gamma$ significantly affects the model's performance.

For Base, we tune the L2 weight decay to $1\times10^{-4}$ following the classic setting of~\cite{devlin2019bert}. Note that we find with experiments that $0.01$ weight decay for this model degrades performance significantly.

\vspace{-3mm}
\subsection{Experiment Results}
\label{sec:experiment_results}

\subsubsection{Comparison with Baselines} Table {\color{red}\ref{tab:major_results}} shows the performance of our model as well as the baselines on the benchmarks. The proposed method CLIM can outperform the previous state-of-the-art method BERT-DAAT by 1.81\% on average. Relying on the strong ability of BERT to extract universal high-quality feature, Base beats the two previous models by a large margin. DANN can get even better results with the assist of adversarial training. Despite the high baseline accuracy given by Base and DANN, our model CLIM outperforms them by 1.00\% and 0.40\% on average, respectively. It's important to note that, for Base and DANN, we fully utilize the ability of BERT and adversarial training by finding appropriate hyperparameters, making them very strong baselines that can even ourperform BERT-DAAT.
Given this, on the two experiments involving airlines dataset, our method CLIM can still obtain higher accuracy than the strong baselines Base and DANN.

The outstanding performance of CLIM on the benchmarks demonstrates its ability to extract discriminative features that best support the final prediction, and to allow the same classifier to be optimal across domains.

\begin{table}[t]
    \centering
    \resizebox{0.48\textwidth}{!}{
    \begin{tabular}{c | c c |c c c | c}
    \toprule[1.5pt]
         \multirow{2}{*}{S$\rightarrow$ T} & \multicolumn{2}{c|}{Previous Models} & \multicolumn{4}{c}{BERT}\\
         \cline{2-7} & PBLM & IATN & Base & DANN & DAAT$^*$  & CLIM\\\midrule[1pt]
         B$\to$ E & 77.60 & 86.50 & 90.50 & 91.67 & 89.57 &  \textbf{92.21}  \\
         K$\to$ B & 74.20 & 84.70 & 88.50 & 89.38 & 87.98 & \textbf{90.13}  \\
         K$\to$ D & 79.90 & 84.10 & 87.90 & 88.89 & 88.81 & \textbf{89.09}  \\
         E$\to$ K & 87.80 & 88.70 & 94.20 & 94.54 & 93.18 & \textbf{94.59} \\
         K$\to$ E & 87.10 & 87.60 & 93.34 & 93.15 & 91.72 & \textbf{93.70} \\
         B$\to$ K & 82.50 & 85.90 & 92.46 & 92.86 & 90.75 & \textbf{93.20} \\ 
         \emph{Average} & 81.52 & 86.25 & 91.15 &  91.75 & 90.34 & \textbf{92.15}   
         \\\midrule[1pt]
         B$\rightarrow$ A & 83.8 &-- & 86.18 &  86.66 & -- & \textbf{87.30}     \\
         A$\rightarrow$ B & 70.6 &-- & 81.65 &  81.50 & -- & \textbf{82.09}      \\    
    \bottomrule[1.5pt]
    \end{tabular}
    }
    \setlength{\abovecaptionskip}{5pt}    
    \caption{Cross-domain sentiment classification accuracy on the Amazon-review dataset and Airlines dataset. B, D, E, K denotes the 4 domains in Amazon-review dataset. A denotes the airlines dataset. \emph{Average} denotes the average accuracy over the domain pairs within the Amazon-review dataset. $^*$The results of DAAT are as reported in their paper~\cite{du2020adversarial}.}
    \label{tab:major_results}
\end{table}

\begin{table}
    \centering
    \begin{tabular}{c|c|c|c} 
    \toprule[1.5pt]
    \multirow{2}{*}{\textbf{MI loss}} & \multirow{2}{*}{\textbf{CL strategy}} & \multicolumn{2}{c}{\textbf{accuracy}} \\
    \cline{3-4}& & B$\to$E & K$\to$D\\
    \midrule[1pt]
    
    & both-domain  & 90.49 & 88.39 \\
    
    \checkmark & both-domain  & 91.98 & 88.44 \\ 
    & in-domain   & 89.78 &  88.99\\ 
    \checkmark &  in-domain  & \textbf{92.21} & \textbf{89.09}\\
    \bottomrule[1.5pt]
    \end{tabular}
\setlength{\abovecaptionskip}{5pt}
\caption{Ablation studies on mutual information maximization and contrastive learning strategy. MI loss: mutual information loss. CL strategy: contrastive learning strategy. In-domain and both-domain denotes in-domain contrastive learning and both-domain contrastive learning respectively.}
\label{tab:ablation}
\end{table}

\vspace{-2mm}
\subsubsection{Ablation Study} Table {\color{red}\ref{tab:ablation}} shows the results of our ablation studies on mutual information maximization and contrastive learning strategy. It illustrates that mutual information maximization consistently improves our model's performance on both domain settings with either contrastive learning strategies. On the other hand, in-domain contrastive learning performs better than both-domain on condition that mutual information loss is also applied, though it sometimes harms performance when mutual information loss is absent. It remains future work to explore when and how in-domain contrastive learning benefits domain adaptation more than both-domain contrastive learning.
\vspace{-5mm}
\section{Conclusion and Broader Impact}
In this paper, we propose CLIM: Contrastive Learning with mutual Information Maximization to promote domain transfer. We are able to outperform strong baselines and the previous state-of-the-art method on two standard benchmarks. Additionally, although back translation works well in this paper, it's still interesting to explore more data augmentation methods and summarize what best serves contrastive learning for cross-domain sentiment classification.



\bibliographystyle{IEEEbib}
\bibliography{strings,refs}

\end{document}